%% file: main.tex
\documentclass{article}

\PassOptionsToPackage{numbers, sort, comma, compress}{natbib}
\usepackage[preprint]{neurips_2020}
\usepackage{comment}

\usepackage[utf8]{inputenc} %
\usepackage[T1]{fontenc}    %
\usepackage{hyperref}       %
\usepackage{url}           \usepackage{booktabs} %
\usepackage{booktabs}       %
\usepackage{amsfonts}       %
\usepackage{nicefrac}       %
\usepackage{microtype}      %
\usepackage{amsmath}
\usepackage{amssymb}
\usepackage{amsthm}
\usepackage{xcolor}
\usepackage{graphicx}       %
\usepackage{subcaption}     %
\usepackage{multirow}

\newcommand{\pname}[1]{{{{I-BERT}}}{#1}}

\title{\pname{}: Inductive Generalization of Transformer to Arbitrary Context Lengths}

\author{%
  Hyoungwook Nam \\
  \texttt{hn5@illinois.edu} \\
  \And
  Seung Byum Seo \\
  \texttt{sbseo2@illinois.edu} \\
  \And
   Vikram Sharma Mailthody \\
  \texttt{vsm2@illinois.edu} \\
  \And
   Noor Michael \\
  \texttt{nsm2@illinois.edu} \\
  \And
   Lan Li \\
  \texttt{lanl2@illinois.edu} \\
    \AND
  \\
  University of Illinois at Urbana-Champaign\\
  Champaign, IL 61820
}

\begin{document}

\maketitle

\input{abstract} 
\input{intro} %
\input{background} %
\input{method} %
\input{model} %
\input{task}
\input{experiment} %
\input{conclusion}

\input{references}
\bibliographystyle{ieeetr}
\bibliography{ref}

\end{document}

%% file: abstract.tex
\begin{abstract}

Self-attention has emerged as a vital component of state-of-the-art sequence-to-sequence models for natural language processing in recent years, brought to the forefront by pre-trained bi-directional Transformer models.
Its effectiveness is partly due to its non-sequential architecture, which promotes scalability and parallelism but limits the model to inputs of a bounded length. 
In particular, such architectures perform poorly on algorithmic tasks, where the model must learn a procedure which generalizes to input lengths unseen in training, a capability we refer to as \textit{inductive generalization}.
Identifying the computational limits of existing self-attention mechanisms, we propose \pname, a bi-directional Transformer that replaces positional encodings with a recurrent layer.
The model inductively generalizes on a variety of algorithmic tasks where state-of-the-art Transformer models fail to do so.
We also test our method on masked language modeling tasks where training and validation sets are partitioned to verify inductive generalization.
Out of three algorithmic and two natural language inductive generalization tasks, \pname{} achieves state-of-the-art results on four tasks.

\end{abstract}

%% file: intro.tex
\section{Introduction}

Neural network models based on self-attention, known as Transformers ~\cite{attentionisalluneed, bert,transformer} have proven to be effective at natural language processing (NLP) tasks such as language modeling, translation, question-answering, and summarization. 
Recent state-of-the-art NLP models are based on pre-trained Transformer encoders such as BERT~\cite{bert}, GPT-2~\cite{gpt2}, XLNet~\cite{xlnet}, RoBERTa~\cite{roberta}, and Megatron-LM~\cite{megatronlm}, which are trained by language modeling tasks.
Compared with the recurrent neural network (RNN) based models, Transformer models are easy to parallelize and are scalable, because parallel self-attention with positional encodings replaces sequential recurrent connections~\cite{transformer}.

This parallel nature of self-attention limits the observable context length of a Transformer to a fixed size.
Prior works have shown that Transformer models cannot generalize to inputs of longer lengths in algorithmic tasks that require inductive bias ~\cite{universaltransformer, theorylimit, numbersequenceprediction}.
Transformer~\cite{transformer} relies on the absolute positional encodings for locality, which cannot be generalized to arbitrarily long context lengths. 
Approaches such as Transformer-XL~\cite{tfxl} and XLNet~\cite{xlnet} enable longer context lengths using the relative positional encodings.
Although they achieve state-of-the-art performance in NLP tasks, there is no proof that they can extend learned rules to unobserved longer inputs.

We define \textit{inductive generalization} to verify a model's capability of extending the rules to unobserved context lengths.
Inductive generalization requires a model to extend the rules learned during training to validation dataset of higher-dimensional samples.
Of course, this requires a change in the method one uses to sample training and validation sets. 
Existing interpolation and extrapolation methods evaluate models on data that share the same sample space with the training data.
But for the case of inductive generalization, we need to split the data samples into a lower-dimensional training set and a higher-dimensional validation set based on the sequence length threshold $k$.
The two sets are mutually exclusive in that the lengths of the validation samples cannot be observed during training.
Successful inductive generalization means that the model infers the rule applicable to higher-dimensional spaces by only observing the samples from a lower-dimensional subspace.

Using the BERT~\cite{bert} architecture as a baseline, we first identify the computational limits of self-attention.
BERT uses absolute positional information and constant computation path~\cite{transformer}. 
We hypothesize that these structural limits prevent the BERT model from achieving inductive generalization. 
To resolve these limitations, we propose \textit{\pname}, an \textit{inductive} bi-directional Transformer that replaces the positional encodings with a recurrent layer.
The RNN layer allows an $O(n)$ computational path to determine sequential dependencies between tokens.
This replacement preserves the performance benefits of self-attention while enjoying the flexibility of RNNs.

To evaluate a model's performance on input lengths unseen during training, we use three algorithmic and two masked language modeling inductive generalization tasks.
Algorithmic tasks from number sequence prediction (NSP) problems~\cite{numbersequenceprediction} are ideal for evaluating inductive generalization as they allow flexible tuning of task difficulty via the number of digits.
For masked character-level and word-level language modeling tasks, we partition the Penn Treebank~\cite{penntreebank} dataset into the training and validation sets by their length to test inductive generalization.

We validate \pname{} on three algorithmic and two masked language modeling inductive generalization tasks. 
The results show that \pname{} is the only model capable of generalizing to longer inputs on algorithmic tasks.
For the masked language modeling tasks, \pname{} outperforms the state-of-the-art (SOTA) models on the character-level task and comes close to the performance of SOTA XLNet in the word-level setup. 
Even though \pname{} does not achieve SOTA on the word-level task, it overcomes BERT's limited context length with higher computational throughput compared to XLNet.

The contributions of this work are as follows:
\begin{itemize}
    \item We formally introduce the problem of generalizing a model to unseen input lengths, which we refer to as \textit{inductive generalization}.
    \item We propose \pname{}, an inductive bi-directional Transformer encoder that replaces positional encodings with a recurrent layer to overcome the computational limits of self-attention.
    \item Out of three algorithmic and two natural language inductive generalization tasks, \pname{} achieves state-of-the-art results on four tasks.
\end{itemize}

%% file: background.tex
\section{Background}
\label{sec:background}

\paragraph{Attention Mechanism}
\label{sec:limitselfattn}

Attention mechanisms were introduced as a supplement to encoder-decoder RNN architectures to address the problem of long-term dependencies between source and target tokens.
This serves a purpose similar to token alignments \cite{ibm}, used in statistical machine translation models.
Two varieties, Luong \cite{luong} and Bahndau \cite{bahndau} attention, take different approaches to combine alignment weights and a context vector to generate the next hidden token and/or an output token.
Vaswani et al. \cite{transformer} introduced a layered encoder-decoder architecture that relies only on \textit{self-attention} between layers and attention on the output of the encoder.
It uses scaled dot-product attention to obtain the context vector, where the scores are computed by dot-products of keys and queries.

\paragraph{Auto-encoder Language Model}
\label{sec:autolm}

A Transformer encoder is often sufficient for language understanding and generation tasks \cite{autoencoder}, which are often fine-tuned via transfer learning to solve NLP tasks.
Devlin et al. introduced BERT, which has a similar architecture but is pre-trained to learn bidirectional representations.
Based on the BERT architecture, recent models such as GPT-2 \cite{gpt2} and MegatronLM \cite{megatronlm} have achieved better performance by increasing the number of parameters by an order of magnitude.
On the other hand, models such as Albert \cite{albert} reduce the number of parameters while retaining performance.
XL-Net \cite{xlnet} integrates techniques from Transformer-XL \cite{tfxl} which tries to extend the context length of Transformer models using recurrence and relative positional encoding \cite{relativeposition}, and introduces a novel pre-training objective that permutes the input sequence.
It also incorporates a two-stream attention mechanism, consisting of a content stream identical to self-attention and a query-stream that does not have information about the current position.

\paragraph{Inductive Bias of Neural Network}
The parallel nature of the self-attention mechanism shortens its critical computation path length to $O(1)$, which enables efficient parallelism at the cost of losing the inductive bias.
In particular, the fact that Transformers cannot generalize learned algorithmic rules beyond observed context lengths has been proved theoretically \cite{theorylimit} and empirically \cite{numbersequenceprediction}.
Universal Transformer \cite{universaltransformer} tries to overcome this limit via repetition, but it shows limited generalization on algorithmic tasks.
Besides attention-based methods, many studies have tried to augment neural networks with external memory structures to learn and generalize algorithmic rules \cite{ntm, stackrnn, nueralgpu, dnc}, but they have not been proven to be effective at practical domains like NLP. 

\paragraph{Recurrent Transformers}
Many recent papers have examined the incorporation of recurrence to Transformer models.
One such model is R-Transformer \cite{rtransformer}, which uses a series of \textit{LocalRNNs} where RNNs operate on a local window.
Other similar papers, such as \cite{arn}, add a distinct RNN separate from the Transformer model, piping its outputs to an attention layer.
The Highway Recurrent Transformer model \cite{highway} recurrently applies an attention mechanism to route information from previous tokens to the current one.
Recurrent Transformer models have also been introduced in domains such as computer vision \cite{rtn} and time series analysis \cite{tft} with similar results.
Such models, although incorporating recurrent components, do not enjoy the full benefits of recurrent models and their ability to extend their performance to longer sequences.

%% file: method.tex
\section{Inductive Generalization}
\label{sec:indgen}

\begin{figure}[ht]
  \centering
  \begin{subfigure}[b]{0.25\linewidth}
        \includegraphics[width=\linewidth]{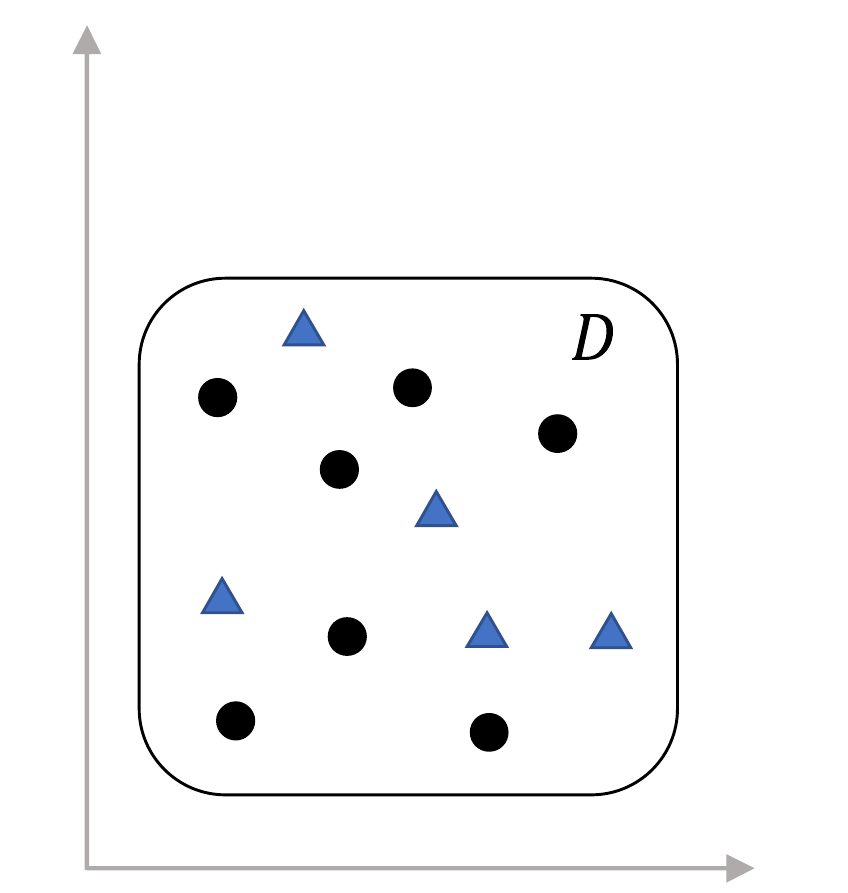}
        \caption{Interpolation}
  \end{subfigure}
  \hfill
  \begin{subfigure}[b]{0.25\linewidth}
        \includegraphics[width=\linewidth]{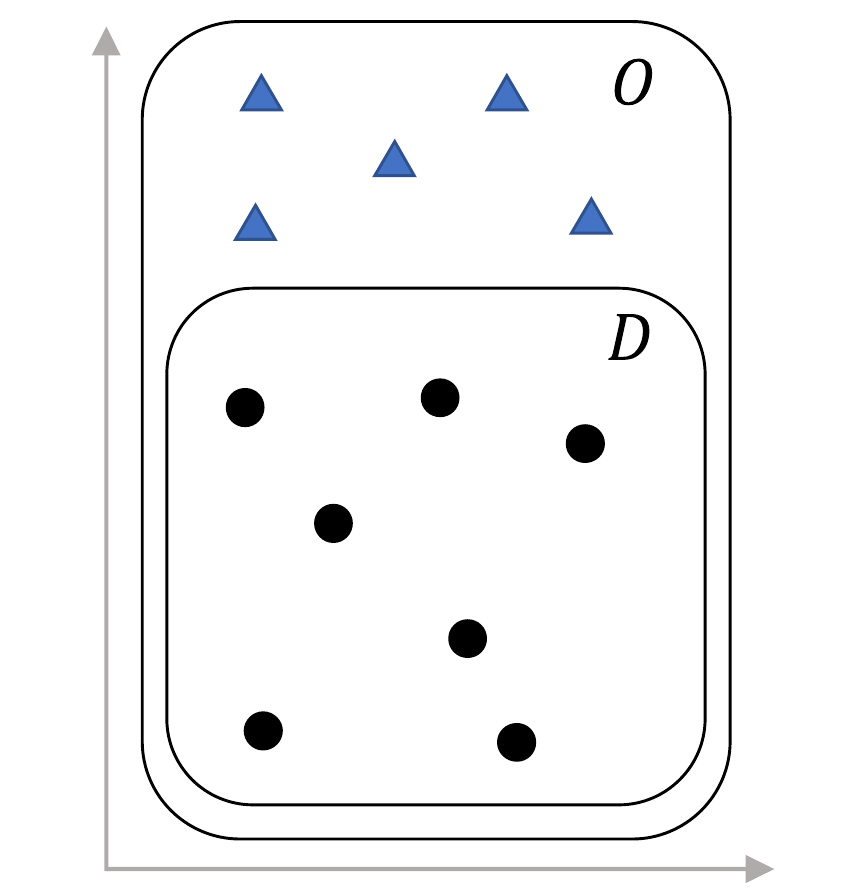}
        \caption{Extrapolation}
  \end{subfigure}
  \hfill
  \begin{subfigure}[b]{0.25\linewidth}
        \includegraphics[width=\linewidth]{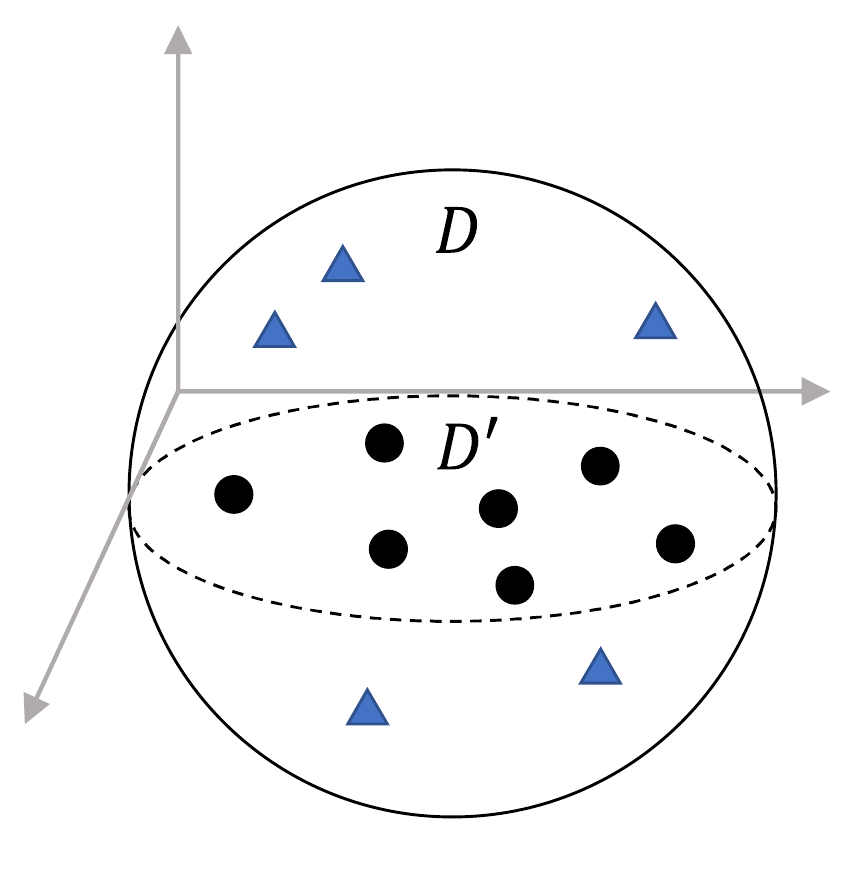}
        \caption{Inductive Generalization}
  \end{subfigure}
  \caption{Conceptual visualizations of how inductive generalization differs from other generalizations. The circles are training data points and the triangles are generalization targets.\vspace{-2ex}}
  \label{fig:indgen}
\end{figure}

Generalization has always been a major issue in machine learning since models easily overfit the training data.
A common practice for resolving this is using a validation dataset independently drawn from the same training data distribution.
Formally, validation and training datasets $\boldsymbol{V},\boldsymbol{T}$ are drawn from the same data distribution $\mathcal{D}$ whose support\footnote{A set of points in the sample space with non-zero probabilities} is $\boldsymbol{D}$.
Since this method mostly verifies \textit{interpolation} (as in Figure~\ref{fig:indgen}a), there have been studies that test \textit{extrapolation} on an out-of-distribution set $\boldsymbol{O}$, of which most of the cases share the \textit{same sample space} as in Figure~\ref{fig:indgen}b.
However, one key aspect of human intelligence is that we can generalize symbolic rules learned from small dimensional samples to unbounded dimensions using induction.
For example, when we learn the addition rule of decimal digits, we first learn the carry rules for adding two or three-digit numbers.
Then, we can easily generalize the rule to longer decimal numbers whose dimensions can be arbitrarily large.
This kind of generalization is also important in natural language processing because the ability to generate arbitrarily long sentences using recursion is a core characteristic of human languages~\cite{recursivemind}.
Therefore, generalization to higher dimensional spaces is important for achieving human-level intelligence, which we define as \textit{inductive generalization}.

The idea of inductive generalization is illustrated in Figure~\ref{fig:indgen}c.
Given a data distribution $\mathcal{D}$, inductive generalization trains a model only by data from $\boldsymbol{D}'$, an intersection of $\boldsymbol{D}$ with a \textit{lower-dimensional subspace}.
To verify if the model can extend the learned rules, we validate it with the higher-dimensional samples from $\boldsymbol{D} \setminus \boldsymbol{D}'$.
Specifically, assume $\mathcal{D}$ has a sample space consists of sequences of tokens whose representation vector space is $\boldsymbol{E}$, so that the sample space is $\boldsymbol{E}^1\cup\boldsymbol{E}^2\dots\cup\boldsymbol{E}^N$.
Once a dataset $\boldsymbol{S}$ is drawn from $\mathcal{D}$, we split the set into the training set $\boldsymbol{T} = \boldsymbol{S} \cap (\boldsymbol{E}^1 \cup \dots \cup \boldsymbol{E}^k)$ and the validation set $\boldsymbol{V} = \boldsymbol{S} \cap (\boldsymbol{E}^{k+1} \cup \dots \cup \boldsymbol{E}^N)$ based on the dimensions of the samples.
The two sets are exclusive in that a validation sample's context length cannot be observed during the training stage.
If a model can generalize the rule learned from $\boldsymbol{T}$ to $\boldsymbol{V}$, we conclude that it is capable of achieving inductive generalization.

Successful inductive generalization means that the model infers the rule applicable to the entire space by only observing the samples from a lower-dimensional subspace.
Hence, the difficulty of inductive generalization depends on the underlying rule of the data distribution.
Parallel architectures like CNNs can inductively generalize if the tokens are independent of one other, while a Turing-complete automaton may be required to emulate the sequential dependency between the tokens.
If the critical path for computing such a dependency is longer than $O(1)$, it is impossible for a Transformer to extend the rule because self-attention only allows constant computational paths~\cite{transformer}.

%% file: model.tex
\section{\pname{} Model}
\label{sec:model}
\begin{figure}[ht]
  \begin{subfigure}[b]{0.32\linewidth}
        \centering
        \includegraphics[width=\linewidth]{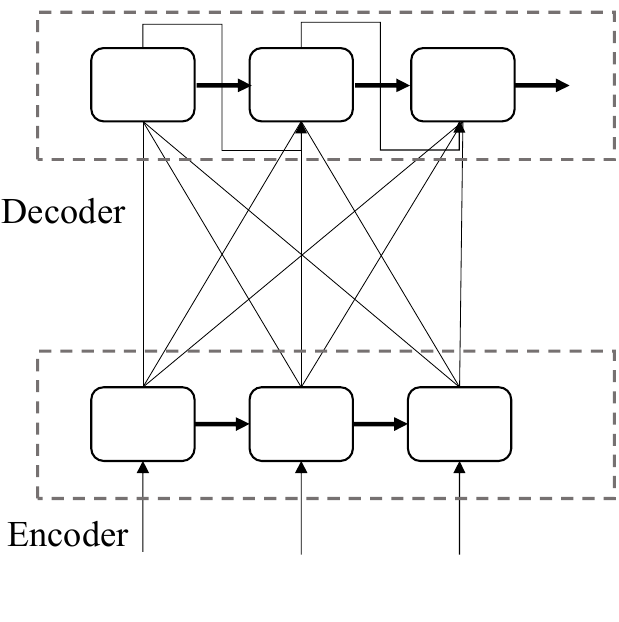}
        \caption{RNN seq2seq}
        \label{fig:seq2seq}
  \end{subfigure}
  \hfill
  \begin{subfigure}[b]{0.32\linewidth}
        \centering
        \includegraphics[width=\linewidth]{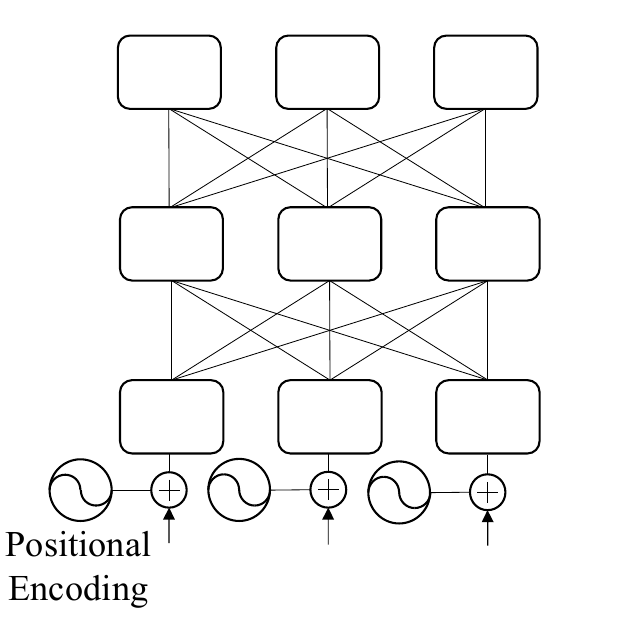}
        \caption{BERT}
        \label{fig:selfatten}
  \end{subfigure}
  \hfill
  \begin{subfigure}[b]{0.32\linewidth}
        \centering
        \includegraphics[width=\linewidth]{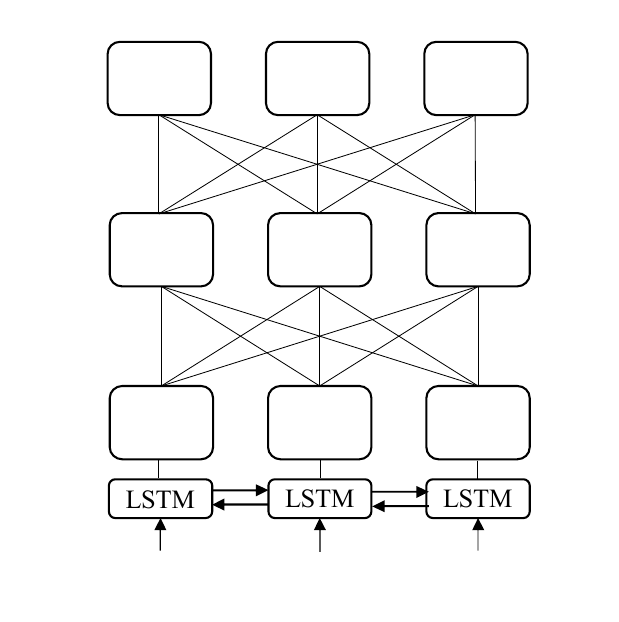}
        \caption{\pname{}}
        \label{fig:ours}
  \end{subfigure}
  \caption{Architecture comparison between (a) RNN sequence-to-sequence model with attention (b) BERT's self-attention and (c) \pname{}'s self-attention with recurrence.}
  \label{fig:compareAttRtx}
\end{figure}

Unlike traditional sequence-to-sequence models as shown in Figure~\ref{fig:seq2seq}, the self-attention mechanism of Transformers shown in Figure~\ref{fig:selfatten} replaces recurrent computations with a positional encoding layer to capture positional relations inside the sequence.
The key benefit of the Transformer architecture is that the computation and gradient paths of each output shortens to $O(1)$ from $O(n)$ of RNNs, making the model easier to stack deeply and achieve parallelism.
However, this makes Transformers incapable of generalizing rules that have a critical path greater than $O(1)$~\cite{theorylimit}. %
For instance, the carry rule of addition has a critical path length of at least $O(log(n))$, making the self-attention based Transformers impossible to learn the rules beyond the observed lengths. 
Furthermore, absolute positional encodings make a model inapplicable to an input longer than any of the training inputs. Relative positional encodings~\cite{relativeposition} can resolve this issue~\cite{tfxl, xlnet}, but they still cannot allow longer computation paths.
Therefore, we propose \pname{}, a BERT based model for inductive generalization.

\pname{} addresses these limitations by augmenting self-attention with a recurrent layer and is shown in the Figure~\ref{fig:ours}.
Besides removing the positional encodings and adding a bi-directional LSTM layer, we share all other architectural details with BERT \cite{bert}.
This architecture brings three upsides.
First, the model can be leveraged for any application where BERT can be used.
Second, both computation and parameter overheads remain constant regardless of the network depth.
Lastly, paths for computing gradients of most of the parameters do not elongate because the recurrent layer is positioned at the bottom.
To sum up, \pname{} overcomes the computational limits of self-attention without sacrificing its adaptability and scalability.

We consider a few variations of \pname{} to verify if our architecture is optimal.
First, since a recurrent layer and a positional encoding layer are not exclusive to one other, we can leverage both methods simultaneously.
However, it is possible that absolute positional encodings cause overfitting.
We can also try using a recurrent layer for every self-attention layer to allow more computational paths.
Our experiment results discussed in Section~\ref{sec:result} show that the only augmenting the self-attention with a recurrent layer such as LSTM is sufficient and effective over its variants. 

%% file: task.tex
\section{Problem and Data}
\label{sec:indgentasks}

\begin{figure}[htbp]
\vspace{-4ex}
\includegraphics[width=\linewidth]{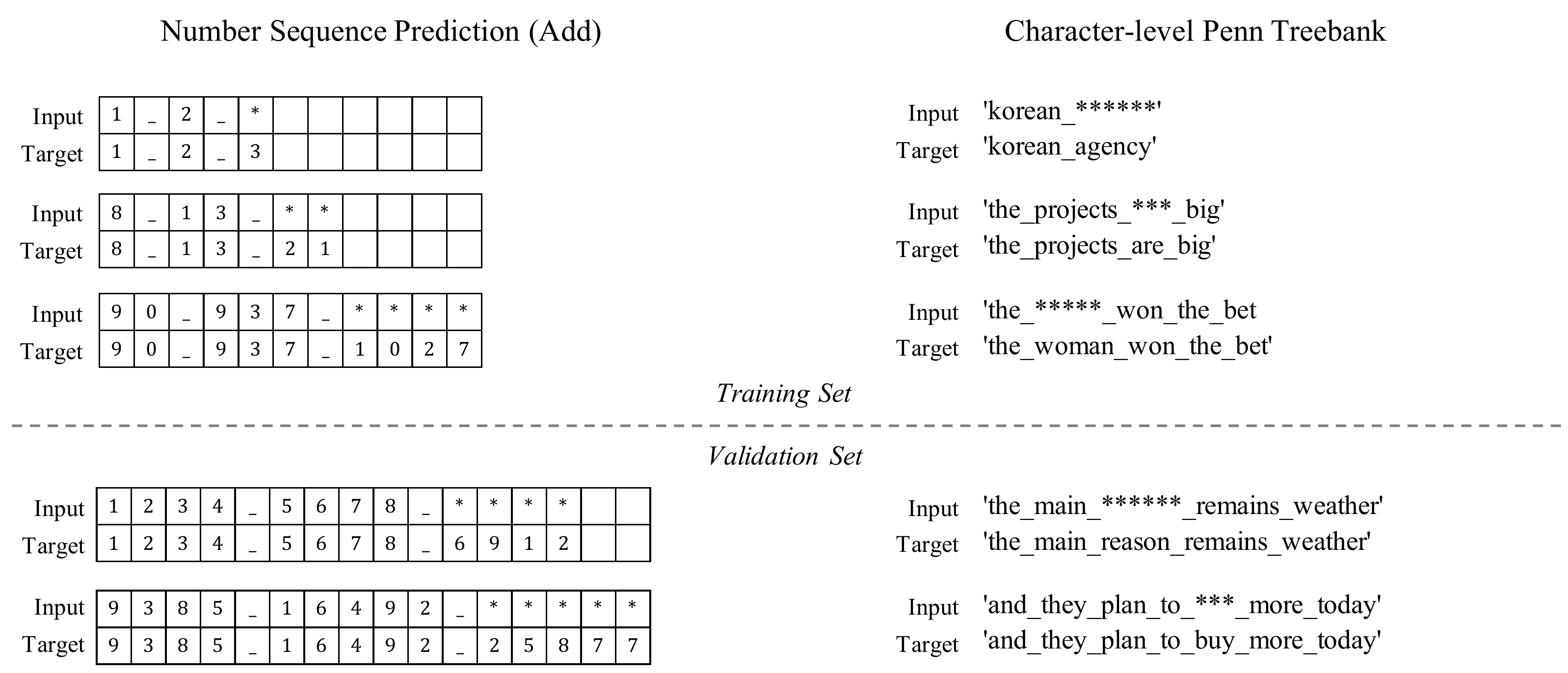}
\caption{Data samples and separation of training and validation datasets. Asterisks represent masks and blank cells represent padding tokens.
}
\vspace{-4ex}
\label{fig_dataset}
\end{figure}

\subsection{Number Sequence Prediction}
Learning arithmetic operations between digital numbers is one of the basic symbolic inference tasks.
Number sequence prediction (NSP) problems \cite{numbersequenceprediction} are well-defined arithmetic rule inference tasks designed for evaluating machine learning models.
One desired property of them is that it is easy to generate inductive generalization validation sets by modifying the number of digits used.
For example, we can easily generate training sets on numbers of length $\leq k$ and evaluation sets on numbers of length $> k$. 
Therefore, we utilize digit-level number sequence prediction tasks from \cite{numbersequenceprediction} for the experiments.
As illustrated in Figure~\ref{fig_dataset}, a digit-level NSP problem is similar to a character-level masked language modeling task in that it requires the model to predict the digits of the masked number given the previous numbers.

We use Fibonacci (Add) and palindrome (Reverse) tasks from \cite{numbersequenceprediction} along with the copying sequence (Copy) tasks in a similar manner.
A data sequence consists of three (two for reverse and copy) numbers $a_1$, $a_2$, and $a_3$, where the initial numbers are sampled randomly and the final number is determined by the generation rule.
Each sequence has a difficulty parameter $d$, which forces all initial numbers to have digits less than or equal to $d$ and makes the second to last number sampled only from $d$-digit numbers.
All the numbers are represented by decimal digits in little-endian order, separated by delimiter tokens.
Shorter sequences are padded with padding tokens to match the length of the longest sequence for mini-batch training.
In an input sequence, all digits of the last number are masked out, and we train the model to restore the original sequence.
To validate inductive generalization, the training dataset has 25600 sequences with $d=2 \dots 12$,and the validation set has 1536 sequences with $d=13 \dots 16$ where the occurrences of difficulties are equally distributed.

\subsection{Penn Treebank Masked Language Modeling}

To verify the generality of \pname{}, we compare it to other state-of-the-art models in masked language modeling tasks.
We use the Penn Treebank dataset \cite{penntreebank} for both the character-level and word-level setups.
The right part in Figure~\ref{fig_dataset} shows how the sentences are sampled and masked for the character-level setup.
In the character-level setup, we mask all characters of a randomly chosen word just like the NSP setup.
The only difference for the word-level setup is that the sentences are tokenized by words, not by characters.
To test inductive generalization, we partition training sentences by their length into separate training and validation datasets.
For the character-level setup, we choose 16 to 192 character sentences for the training dataset and 193 to 224 character sentences for the validation dataset.
Likewise, we choose 2 to 32 word sentences for training, and 33 to 64 word sentences for validation in the word-level setup.
The thresholds are chosen to make the train-to-validation size ratio around 4:1.
All sequences are padded to match the length of the longest sentence in the dataset.

%% file: experiment.tex
\section{Experimental Result}
\label{sec:result}

\subsection{Experimental Setup}

\renewcommand{\arraystretch}{1.3}
\begin{table}[htbp]
    \vspace{-3ex}
    \caption{Baseline and \pname{} Model hyperparameters and parameter sizes used for experiments.}
    \label{tab:modelparam}
    \vspace{1ex}
    \centering
    \begin{tabular}{ lllll } \toprule
     Model & \textbf{\#}Layers  & \textbf{\#}Heads & Hidden Size & Param Size\\ \midrule
    
    \pname{} (ours)    & 12   & 12  & 768 & 69M\\
     BERT~\cite{bert}   & 12  & 12 & 768 & 71M  \\
     XLNet~\cite{xlnet}     & 12  & 12 & 768 & 73M   \\
     LSTM seq2seq & 2   &1   &768   &   8M  \\ \bottomrule
     \hline
    \end{tabular}
    \vspace{-3ex}
\end{table}

\paragraph{Baseline models }
Since we follow a BERT-like architecture, we choose BERT-base and XLNet-base as baseline models for comparison. 
Table~\ref{tab:modelparam} describes the hyperparameters used in the evaluation. 
The parameter size overhead is marginal compared to the rest of the models as we only add a recurrent layer at the bottom.
We also compare with the LSTM seq2seq model with attention from the fairseq~\cite{fairseq} library as a baseline to show that our method outperforms traditional combinations of LSTM and attention.
The pre-trained models cannot be applied to our tasks, so we replicate the existing PyTorch implementations~\cite{huggingface} and train them from scratch.
Our implementations are publicly available on GitHub~\footnote{https://github.com/hwnam831/ibert}.
  
\paragraph{Controlled variables}
 
We train the baseline and \pname{} models using an Adam optimizer with the learning rate decay of  0.97 per epoch.
For NSP tasks, we train them for 50 epochs with an initial learning rate of 3e-5, and for both character-level and word-level masked LM tasks, we train them for 100 epochs with an initial learning rate of 1e-4.
To fit our NVIDIA Titan V GPU used for training, the batch sizes are set to 16 in character-level masked LM tasks, and 32 for the other tasks.

\subsection{Number Sequence Prediction Result}

\begin{figure}[htbp] 
\vspace{-2ex}
  \begin{subfigure}[b]{0.33\linewidth}
    \centering
    \includegraphics[width=1.0\linewidth]{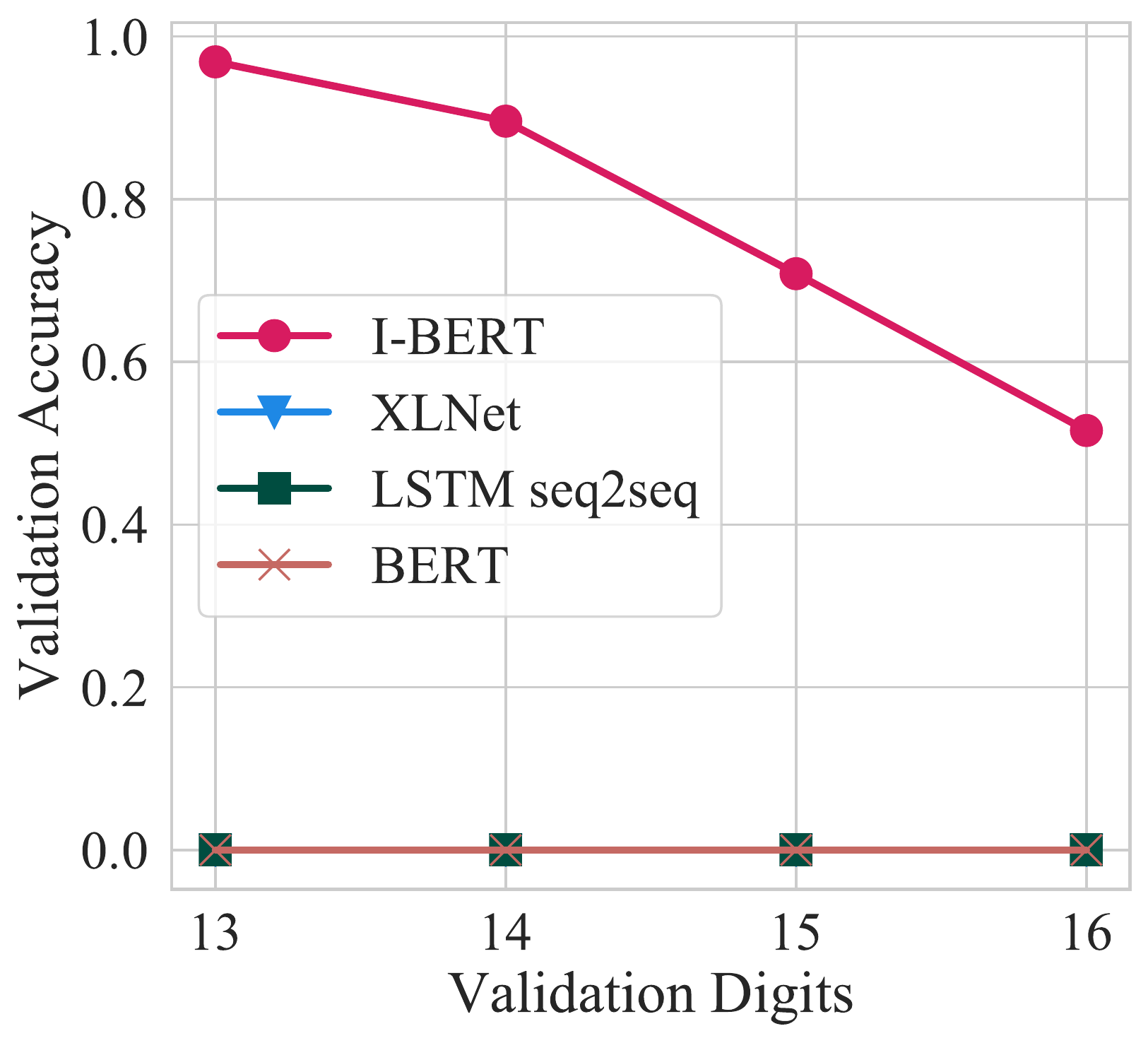} 
    \caption{Addition} 
    \label{fig:nsp_a} 
  \end{subfigure}%
  \begin{subfigure}[b]{0.33\linewidth}
    \centering
    \includegraphics[width=1.0\linewidth]{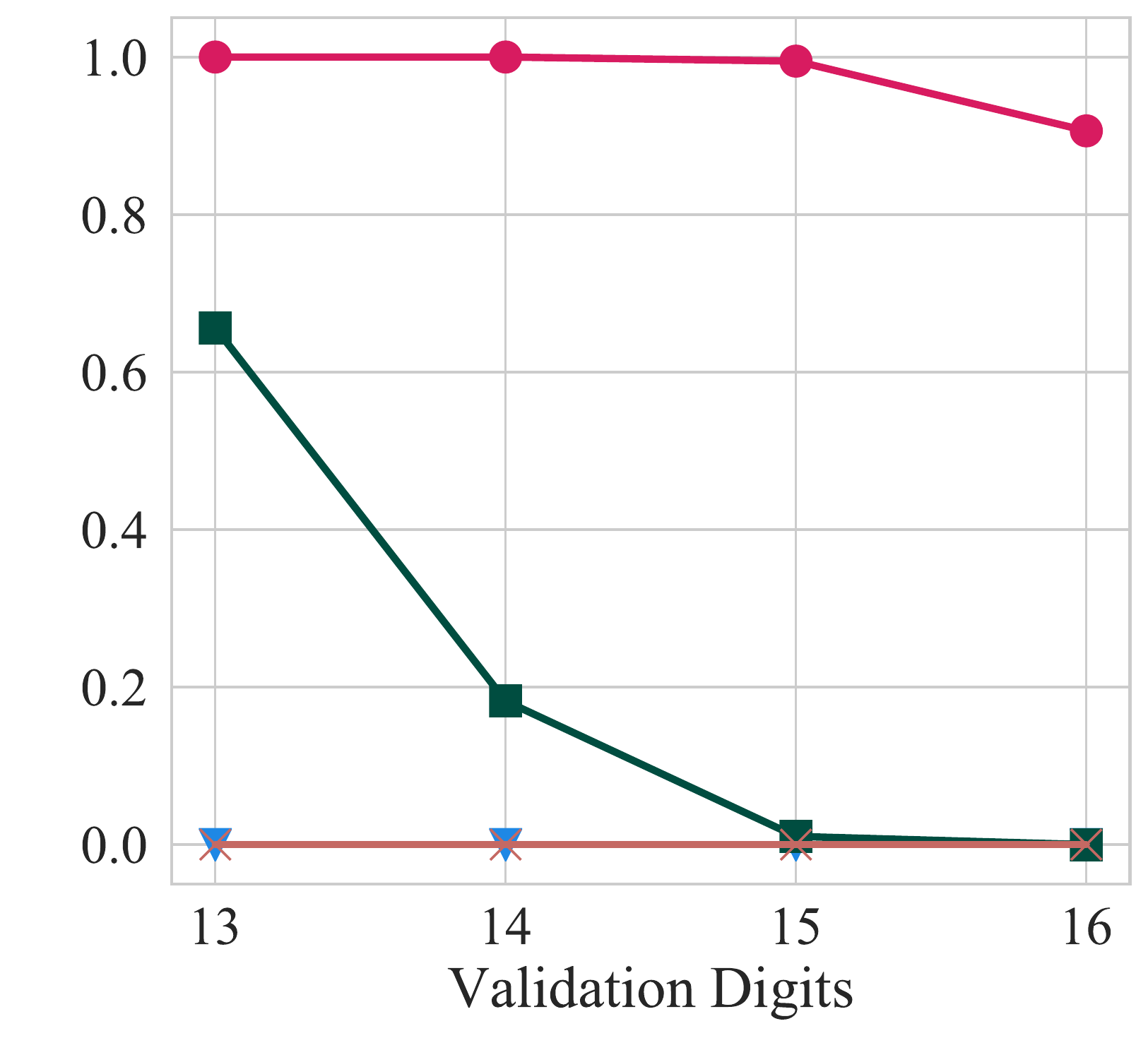} 
    \caption{Copy} 
    \label{fig:nsp_b} 
  \end{subfigure}%
  \begin{subfigure}[b]{0.33\linewidth}
    \centering
    \includegraphics[width=1.0\linewidth]{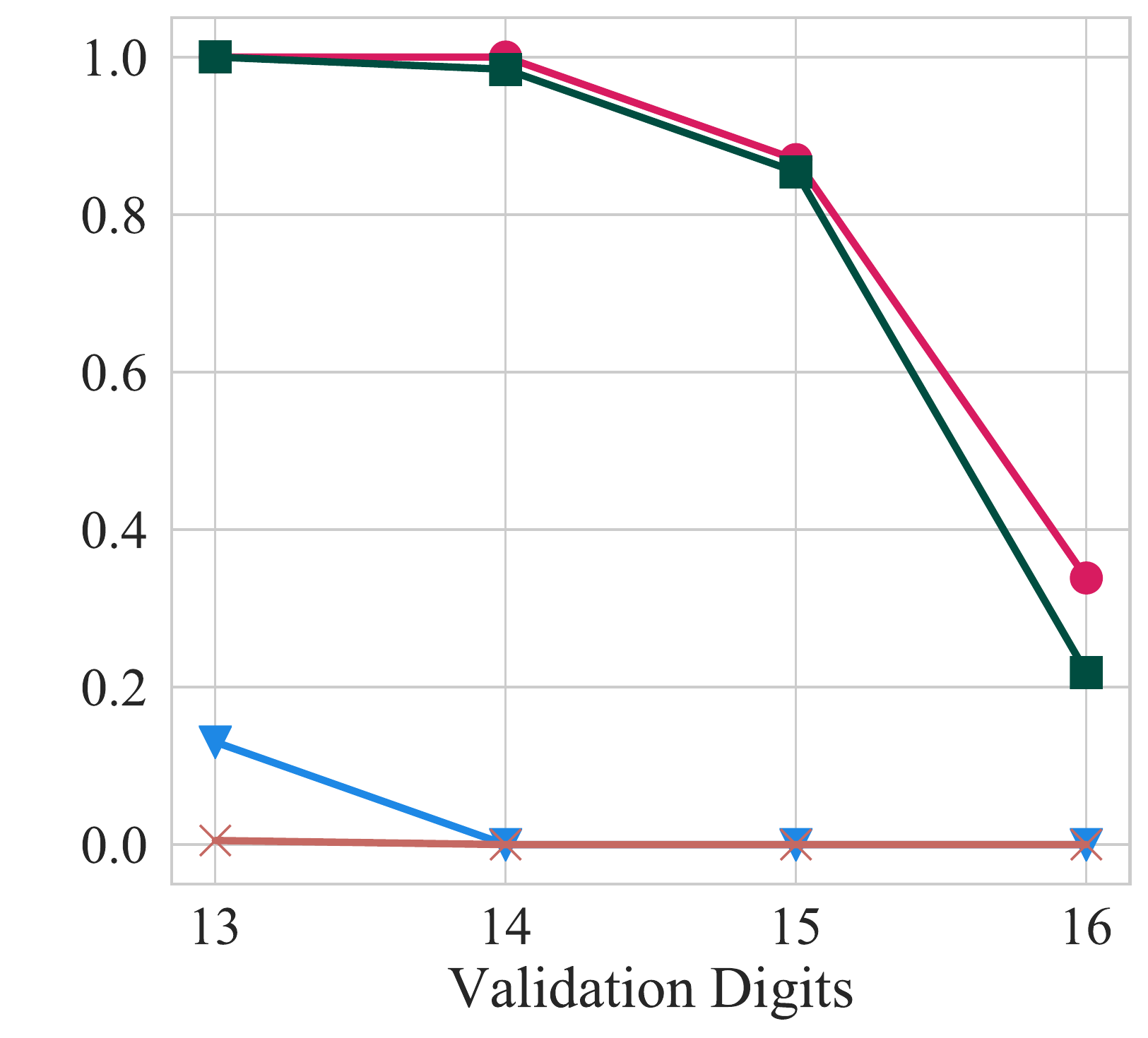} 
    \caption{Reverse} 
    \label{fig:nsp_c} 
  \end{subfigure} 
  \caption{Across all the different NSP tasks, \pname{} achieves SOTA accuracy.}
  \label{fig:nsp_experiment} 
  \vspace{-2ex}
\end{figure}

The objective of the number sequence prediction experiment is to see if a machine learning model can extend learned algorithmic rules to unseen long inputs.
Figure~\ref{fig:nsp_experiment} shows validation sequence accuracy of the models when trained on sequences up to $k=12$ digits.
Although all models could achieve near 100\% sequence accuracy during training, most of them find it hard to generalize on out-of-distribution validation sets.
As shown in Figure~\ref{fig:nsp_a}, \pname{} is the only model that successfully extends the addition rule to longer input sequences, while other state-of-the-art Transformers completely fail to do so.
The result is expected given the nature of the carry rule in addition.
Because carry digits require at least $O(log(n))$ steps to be determined, it is impossible for typical Transformers with $O(1)$ compute paths to determine them.
Therefore, the only possible way for them to train is to memorize the specific rule for each observed length, which cannot be generalized to arbitrarily long inputs.

A similar pattern appears on simpler algorithmic tasks such as Copy and Reverse as seen in Figure~\ref{fig:nsp_b} and Figure~\ref{fig:nsp_c}.
The failures of BERT are expected because the encoded positions of longer parts of validation sequences cannot be observed during training.
The interesting result is that XLNet's relative positional encoding also finds it hard to generalize on validation sets, despite the claim that it can extend the Transformer's context lengths.
This can be explained by temporal distances between corresponding digits.
The relative temporal distance between corresponding digits becomes longer as we use longer digit sequences, but the relative positional encodings cannot resolve this if such long distances had not been observed during training.
On the other hand, \pname{}'s recurrent layer can encode such relative position information within a sequence as shown in Figure~\ref{fig:nsp_experiment}.

Comparisons with the LSTM seq2seq model proves that our architecture which combines self attention with an RNN is superior to the traditional combination of RNN and attention.
The LSTM seq2seq model fails to generalize on the addition task and, although it can achieve some generalization on the other tasks, its validation accuracy degrades faster than that of \pname{} for longer inputs.
Therefore, compared to the single-head content-based attention, we can conclude that the key-query-value mechanism and multi-head architecture of self-attention are beneficial for inductive generalization.

\begin{figure}[h!] 
\vspace{-2ex}
  \begin{subfigure}[b]{0.33\linewidth}
    \centering
    \includegraphics[width=1.0\linewidth]{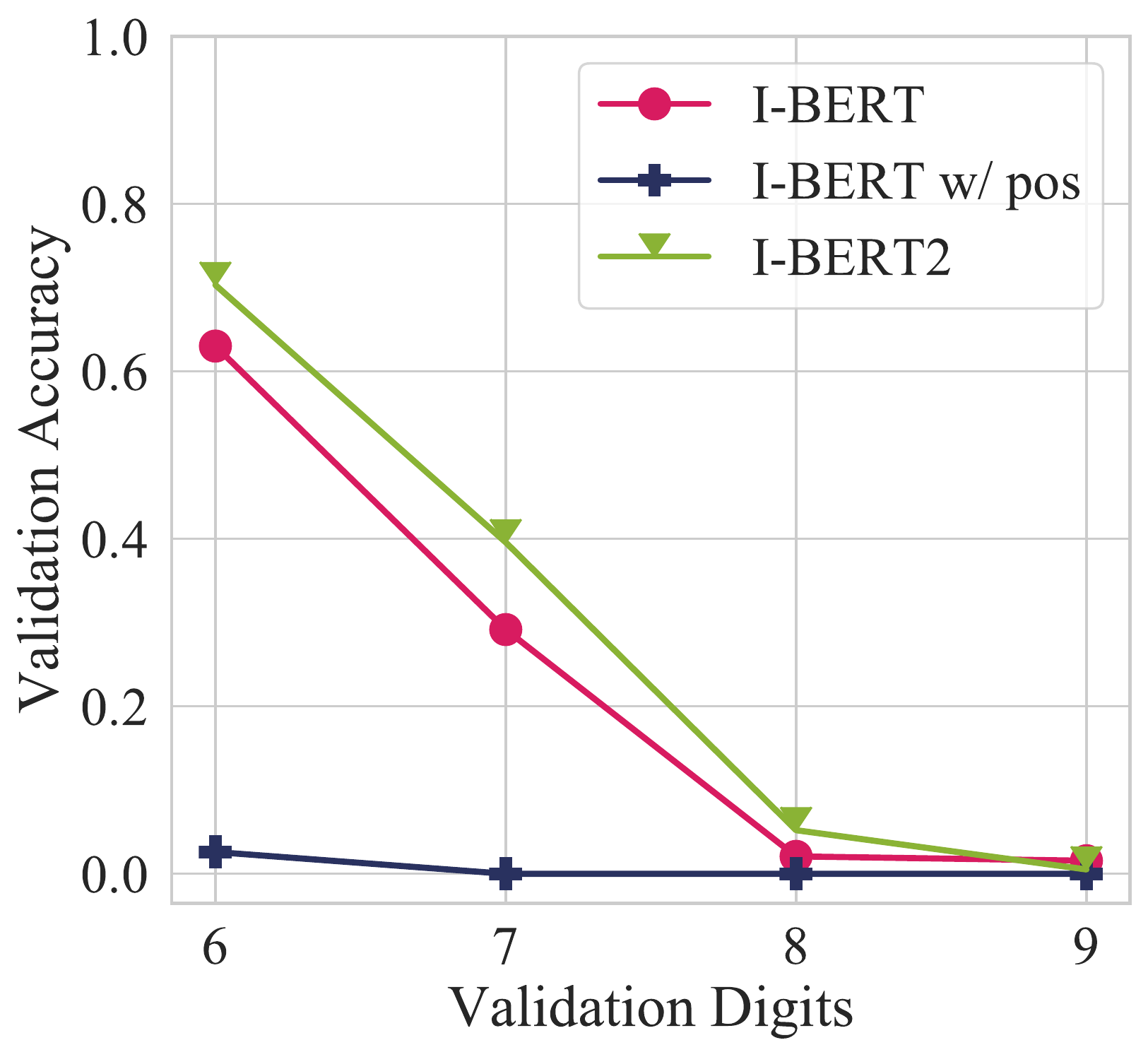} 
    \caption{$k=2\dots5$} 
    \label{fig:ablation_a} 
  \end{subfigure}%
  \begin{subfigure}[b]{0.33\linewidth}
    \centering
    \includegraphics[width=1.0\linewidth]{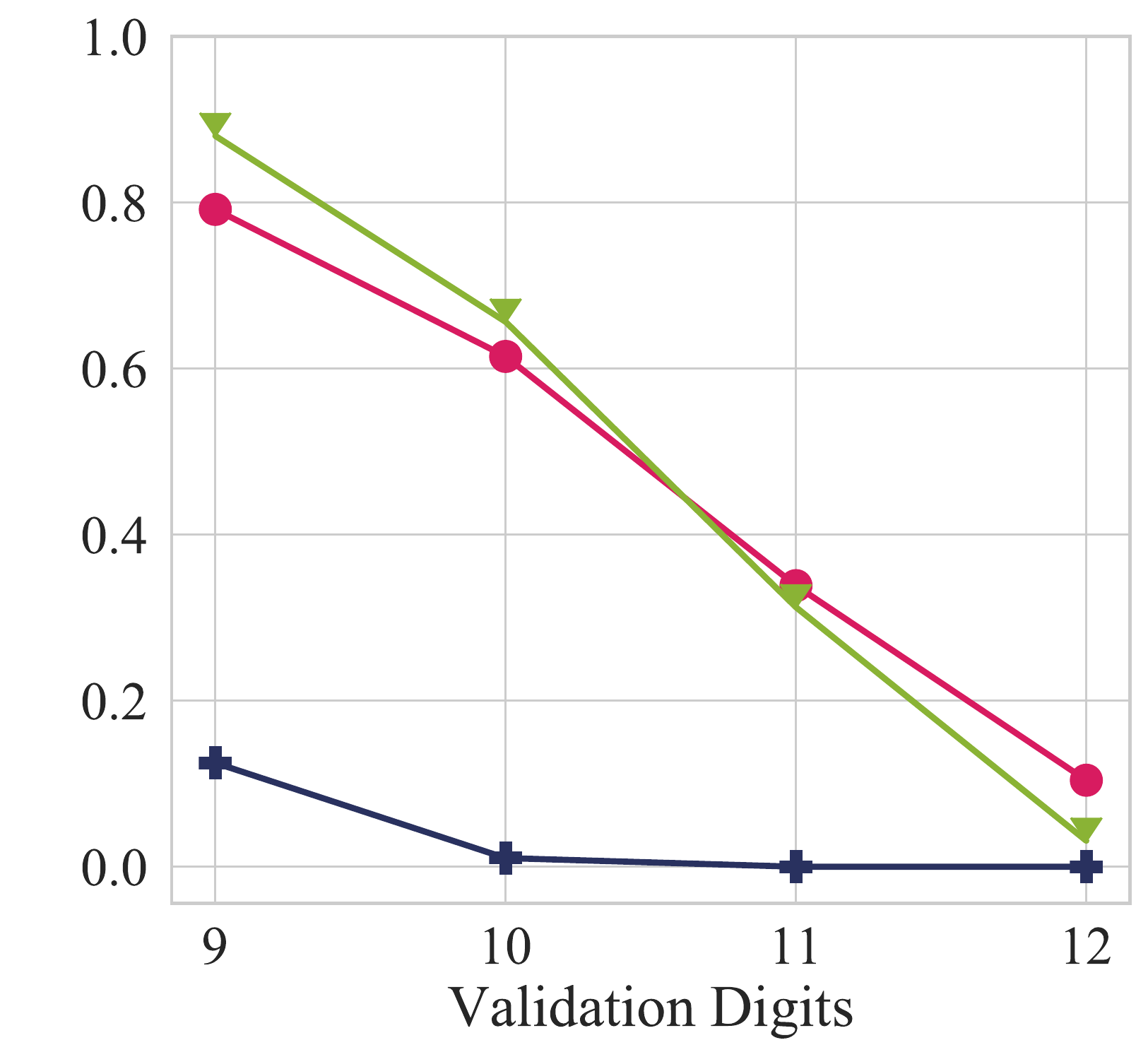} 
    \caption{$k=2\dots8$} 
    \label{fig:ablation_b} 
  \end{subfigure} 
  \centering
  \begin{subfigure}[b]{0.33\linewidth}
    \centering
    \includegraphics[width=1.0\linewidth]{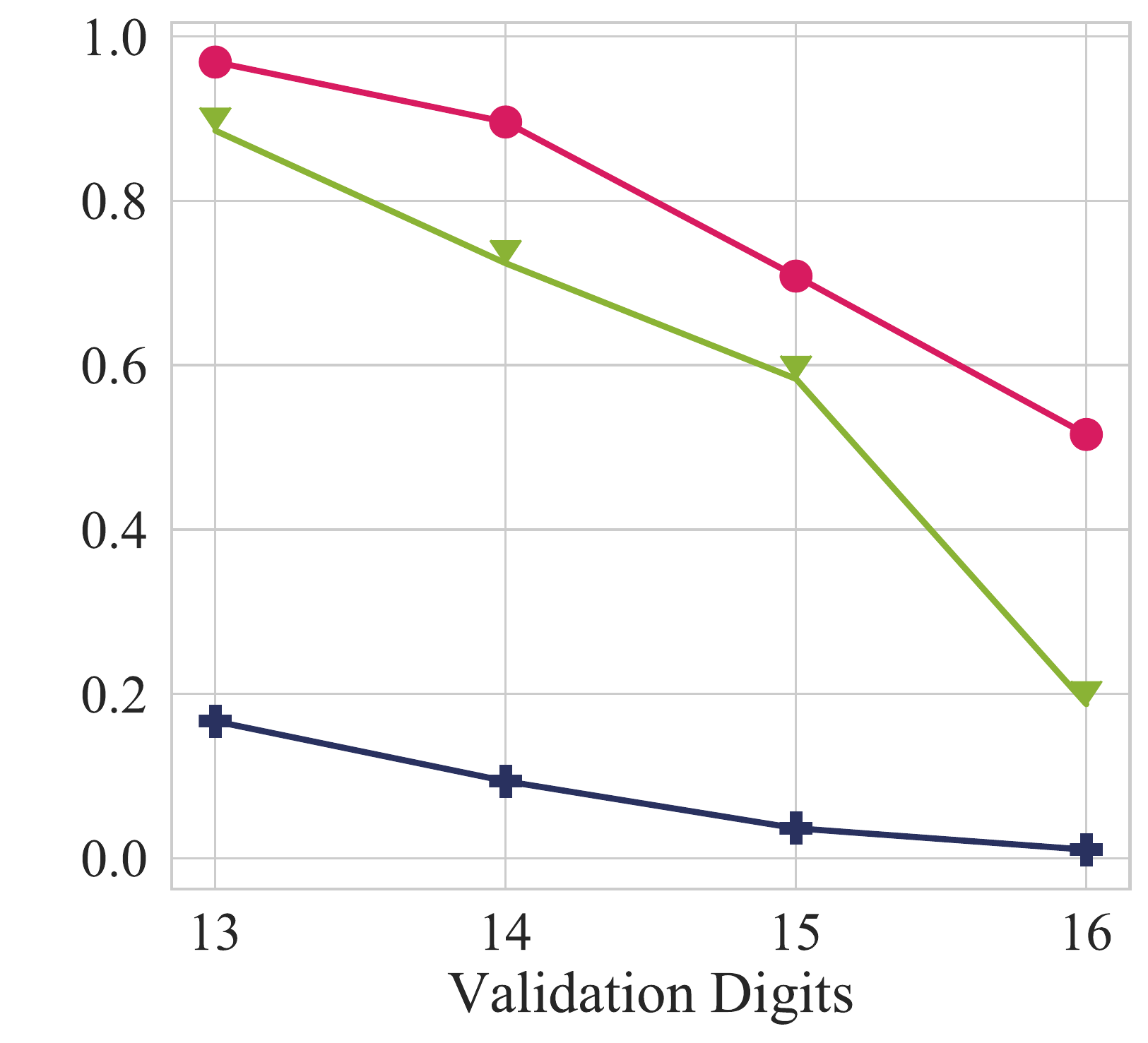} 
    \caption{$k=2\dots12$} 
    \label{fig:ablation_c} 
  \end{subfigure}%
  \caption{Ablation study of \pname{} using addition tasks in different training and validation setups. $k=i\dots j$ means that the difficulties of the training set ranges from $i$ to $j$. \pname{}2 is an \pname{} model where every feed-forward layer is replaced by an LSTM layer.}
  \label{fig:nsp_ablation} 
  \vspace{-2ex}
\end{figure}

We perform an ablation study to find the optimal design variant for \pname{}.
Our ablation study covers two variants: (1) the combination with the positional encodings, and (2) replacement of every feed-forward layers with an LSTM layer.
We perform the ablation study using NSP addition tasks and the results are shown in Figure~\ref{fig:nsp_ablation}.
As before, all models can achieve 100\% sequence accuracy during training.
From our study, we hypothesized that providing absolute position information to the architecture makes it harder to inductively generalize.
These results empirically demonstrate this hypothesis in that the addition of the positional encodings makes a model overfit to the training data.
Moreover, adding additional recurrent layers as in the case of \pname{}2 is not advantageous. %
Figure~\ref{fig:ablation_c} shows that \pname{} generalizes better if we provide longer sequences during training.
In contrast, \pname{}2, an \pname{} where all feed-forward layers are replaced by LSTM layers, shows an opposite trend.
It could not generalize well given longer training sequences, indicating that adding more recurrence makes the model weaker to longer sequence inputs.
Therefore, the proposed architecture is best performing implementation of \pname{} among the different variants we studied.

\subsection{Penn Treebank Masked Language Modeling}

\renewcommand{\arraystretch}{1.3}
\begin{table}[htbp]
\caption{Penn Treebank masked language model results. The models are evaluated by the metrics of training sequences per second (Seq/s), bits per character (BPC), and perplexity (PPL).}
\vspace{2ex}
\label{table:maskedlm}
\centering
\begin{tabular}{lllllllllll}  \toprule
 {}   &   \multicolumn{5}{c}{Character-level} & \multicolumn{5}{c}{Word-level} \\ \cmidrule(lr){2-6} \cmidrule(lr){7-11} 
\multirow{2}{*}{Models} & \multirow{2}{*}{Seq/s} & \multicolumn{2}{c}{Training} & \multicolumn{2}{c}{Validation} & \multirow{2}{*}{Seq/s} & \multicolumn{2}{c}{Training} & \multicolumn{2}{c}{Validation} \\
\cmidrule(lr){3-4} \cmidrule(lr){5-6} \cmidrule(lr){8-9} \cmidrule(lr){10-11} 
{}                      & {}        & Acc & BPC     & Acc  & BPC & {} & Acc & PPL & Acc & PPL \\  \midrule
   \textbf{\pname{}} (ours)    & 72 & 0.44 & 1.46       & \textbf{0.32}      & \textbf{2.07} & 343      & 0.58 & 6.86    & 0.32  & 21.02 \\ 
    XLNet \cite{xlnet}  & 35        & 0.36 & 1.81   & 0.30  & 2.10 &  217      & 0.58 & 5.22    & \textbf{0.39}  & \textbf{14.05}\\
    LSTM seq2seq        & 200       & 0.35 & 1.97   & 0.30  & 2.16 & 1280      & 0.34 & 15.11   & 0.29 & 22.95 \\
    BERT \cite{bert}    & 88        & 0.10 & 2.69   & 0.08  & 2.77 &  394      & 0.57 & 5.31    & 0.25 & 34.43 \\ \bottomrule
\end{tabular}

\end{table}

The masked language modeling results in Table~\ref{table:maskedlm} show that \pname{} can be extended on the NLP tasks.
\pname{} outperforms both XLNet and BERT on the character-level task, while its performance lies between XLNet and BERT on the word-level task.
Considering that word-level tokenization makes the natural language sentences much shorter than character-level tokenization, it seems natural that inductive generalization does not play a key role in word-level tasks.
However, our architecture outperforms BERT in both tasks because BERT cannot handle masked tokens beyond the observed context lengths during training.
Since the performance overhead of our model is much smaller than the expensive two-stream attention and relative positional encoding of XLNet, our method is a cheaper alternative to overcome the limited context length of basic Transformers.

%% file: conclusion.tex
\section{Discussion}

In this section, we discuss possible issues that arise in the design and application of our model.
The first possible issue is that even though we extend the computation path of the Transformer architecture from $O(1)$, it is still limited to $O(n)$.
In fact, the computation path depends on the number of \textit{timesteps} $t$, which is proportional to the number of tokens $n$ in our sequence-to-sequence setup because we only allow one timestep per token.
If we can find a training scheme that allows multiple timesteps per token, the same architecture can be applied to tasks of any time complexity.
For instance, if we use our architecture along with reinforcement learning, it is possible to let the model take enough time to produce each output by allowing \textit{no-op} actions.

Another issue is that the model's inductive generalization accuracy degrades as we give longer input sequences.
So one can argue that the model does not truly inductively generalize because a human's inductive generalization capability does not degrade for longer context.
However, this is not a fair comparison because we need to \textit{write down} the symbols somewhere to extend the rules to arbitrary lengths.
If we solely rely on our memory to perform such generalization, our performance would degrade much faster than what we observe from \pname{}.
That is, we need to devise a reliable external memorization method, like memory-augmented neural networks (MANN), to achieve inductive generalization that does not degrade over longer contexts.

The final issue is that \pname{} outperforms the others only in character-level tasks so the practicality of our model is questionable.
However, advancements in character-level tasks~\cite{character-level-transformer} are possibly more important than improvements in word-level tasks.
Since word embeddings require a predetermined vocabulary, word-level models have a limited scope of applicability~\cite{liang2017combining}.
For example, it is impossible for a word-level model to process typoglycemia sentences like "Aoccdrnig to a rseearch taem at Cmabrigde" because the words will be translated to \textit{unknown} tokens.
Therefore, advancements in character-level tasks will greatly extend the scope of NLP applications of our model.

\section{Conclusion}

In this paper, we examined the generalization of Transformer models to context lengths unseen in training, which we refer to as inductive generalization.
We identified the limits of self-attention mechanisms to this problem, in turn evaluating a series of other common model architectures such as LSTM, BERT, and XLNet.
We proposed \pname{}, a novel Transformer architecture which replaces the positional encoding with a recurrent layer to overcome computational limits of self-attention.
Also, we defined inductive generalization task setups of number sequence prediction tasks and Penn Treebank masked language modeling tasks to check if a model truly extends its context length.
\pname{} outperforms state-of-the-art Transformers at algorithmic tasks and character-level language modeling tasks.
We demonstrated that \pname{} is capable of generalizing to unseen sequence lengths while maintaining a competitive space and time complexity.

\section*{Broader Impact}

Inductive generalization is one of the key challenges in implementing human-level artificial general intelligence.
For example, we are naturally capable of extending the Addition rule to arbitrarily long digits after learning rules for limited digit additions.
If we can make a breakthrough for achieving inductive generalization via deep learning, the finding can shed light on other fields studying human intelligence, including neuroscience, cognitive psychology, and linguistics.

These findings will significantly improve the understanding of inductive bias and further facilitate communication between two different groups and mitigate the social inequality caused by artificial intelligence. In particular, from an education perspective, it will provide quantifiable insights for educators to understand to what extent of knowledge students would be able to produce when given a certain amount of content. From an industry perspective, leaders and technicians will experience a clear agreement process by being provided with predictions and estimates automatically generated from inductive bias. 

Unlike the number prediction dataset, however, any natural language dataset carries a challenge where it fails to include interests and conflicts from diverse societal representatives. In our setting, we took advantage of the number prediction dataset, which is relatively free from demographic bias \cite{socialimpact}. Since more investigation is needed to overcome demographic bias, we will keep exploring the difference between the number prediction dataset and the natural language dataset to find an objective model which most members of society can intuitively accept.

\begin{ack}
We would like to thank Julia Hockenmaier, Zhenbang Wang and our advisors Josep Torrellas, Wen-mei Hwu, Bertram Lud\"ascher and other anonymous reviewers for their helpful comments and feedbacks. 
Vikram Sharma Mailthody was partly supported by Joan and Lalit Bahl Fellowship from University of Illinois. 
Lastly, this work would not have been possible without the generous hardware donations from NVIDIA, AMD and the University of Illinois. 

\end{ack}

%% file: main.bbl
\begin{thebibliography}{10}

\bibitem{attentionisalluneed}
A.~Vaswani, N.~Shazeer, N.~Parmar, J.~Uszkoreit, L.~Jones, A.~N. Gomez, L.~u.
  Kaiser, and I.~Polosukhin, ``{A}ttention is {A}ll you {N}eed,'' in {\em
  Advances in Neural Information Processing Systems 30} (I.~Guyon, U.~V.
  Luxburg, S.~Bengio, H.~Wallach, R.~Fergus, S.~Vishwanathan, and R.~Garnett,
  eds.), pp.~5998--6008, Curran Associates, Inc., 2017.

\bibitem{bert}
J.~Devlin, M.-W. Chang, K.~Lee, and K.~Toutanova, ``{BERT}: {P}re-training of
  {D}eep {B}idirectional {T}ransformers for {L}anguage {U}nderstanding,'' in
  {\em Proceedings of the 2019 Conference of the North {A}merican Chapter of
  the Association for Computational Linguistics: Human Language Technologies,
  Volume 1 (Long and Short Papers)}, (Minneapolis, Minnesota), Association for
  Computational Linguistics, June 2019.

\bibitem{transformer}
I.~Sutskever, O.~Vinyals, and Q.~V. Le, ``Sequence to {S}equence {L}earning
  with {N}eural {N}etworks,'' in {\em Proceedings of the 27th International
  Conference on Neural Information Processing Systems - Volume 2}, NIPS’14,
  (Cambridge, MA, USA), p.~3104–3112, MIT Press, 2014.

\bibitem{gpt2}
A.~Radford, J.~Wu, R.~Child, D.~Luan, D.~Amodei, and I.~Sutskever, ``Language
  models are unsupervised multitask learners,'' 2018.

\bibitem{xlnet}
Z.~Yang, Z.~Dai, Y.~Yang, J.~Carbonell, R.~R. Salakhutdinov, and Q.~V. Le,
  ``{XLNet: Generalized Autoregressive Pretraining for Language
  Understanding},'' in {\em Advances in Neural Information Processing Systems
  32} (H.~Wallach, H.~Larochelle, A.~Beygelzimer, F.~d~Alch\'{e}-Buc, E.~Fox,
  and R.~Garnett, eds.), pp.~5753--5763, Curran Associates, Inc., 2019.

\bibitem{roberta}
Y.~Liu, M.~Ott, N.~Goyal, J.~Du, M.~Joshi, D.~Chen, O.~Levy, M.~Lewis,
  L.~Zettlemoyer, and V.~Stoyanov, ``Ro{BERT}a: {A} {R}obustly {O}ptimized
  {BERT} {P}retraining {A}pproach,'' {\em CoRR}, vol.~abs/1907.11692, 2019.

\bibitem{megatronlm}
M.~Shoeybi, M.~Patwary, R.~Puri, P.~LeGresley, J.~Casper, and B.~Catanzaro,
  ``{Megatron-LM: Training Multi-Billion Parameter Language Models Using Model
  Parallelism},'' 2019.

\bibitem{universaltransformer}
M.~Dehghani, S.~Gouws, O.~Vinyals, J.~Uszkoreit, and L.~Kaiser, ``{Universal
  Transformers},'' {\em CoRR}, vol.~abs/1807.03819, 2018.

\bibitem{theorylimit}
M.~Hahn, ``{Theoretical limitations of self-attention in neural sequence
  models},'' {\em Transactions of the Association for Computational
  Linguistics}, vol.~8, pp.~156--171, 2020.

\bibitem{numbersequenceprediction}
H.~Nam, S.~Kim, and K.~Jung, ``{Number Sequence Prediction Problems for
  Evaluating Computational Powers of Neural Networks},'' in {\em Proceedings of
  the AAAI Conference on Artificial Intelligence}, vol.~33, pp.~4626--4633,
  2019.

\bibitem{tfxl}
Z.~Dai, Z.~Yang, Y.~Yang, J.~Carbonell, Q.~Le, and R.~Salakhutdinov,
  ``Transformer-{XL}: {A}ttentive {L}anguage {M}odels beyond a {F}ixed-{L}ength
  {C}ontext,'' in {\em Proceedings of the 57th Annual Meeting of the
  Association for Computational Linguistics}, (Florence, Italy),
  pp.~2978--2988, Association for Computational Linguistics, July 2019.

\bibitem{penntreebank}
A.~Taylor, M.~Marcus, and B.~Santorini, ``{The Penn Treebank: An Overview},''
  in {\em Treebanks}, pp.~5--22, Springer, 2003.

\bibitem{ibm}
P.~F. Brown, S.~A. Della~Pietra, V.~J. Della~Pietra, and R.~L. Mercer, ``{The
  Mathematics of Statistical Machine Translation: Parameter Estimation},'' {\em
  Computational Linguistics}, vol.~19, no.~2, pp.~263--311, 1993.

\bibitem{luong}
T.~Luong, H.~Pham, and C.~D. Manning, ``{Effective Approaches to
  Attention-based Neural Machine Translation},'' in {\em Proceedings of the
  2015 Conference on Empirical Methods in Natural Language Processing},
  (Lisbon, Portugal), pp.~1412--1421, Association for Computational
  Linguistics, Sept. 2015.

\bibitem{bahndau}
D.~Bahdanau, K.~Cho, and Y.~Bengio, ``{Neural Machine Translation by Jointly
  Learning to Align and Translate},'' in {\em 3rd International Conference on
  Learning Representations, {ICLR} 2015, San Diego, CA, USA, May 7-9, 2015,
  Conference Track Proceedings} (Y.~Bengio and Y.~LeCun, eds.), 2015.

\bibitem{autoencoder}
A.~Radford, J.~Wu, R.~Child, D.~Luan, D.~Amodei, and I.~Sutskever, ``{Language
  Models are Unsupervised Multitask Learners},'' {\em OpenAI Blog}, vol.~1,
  no.~8, p.~9, 2019.

\bibitem{albert}
Z.~Lan, M.~Chen, S.~Goodman, K.~Gimpel, P.~Sharma, and R.~Soricut, ``{Albert: A
  lite BERT for Self-supervised Learning of Language Representations},'' {\em
  arXiv preprint arXiv:1909.11942}, 2019.

\bibitem{relativeposition}
P.~Shaw, J.~Uszkoreit, and A.~Vaswani, ``Self-attention with relative position
  representations,'' in {\em Proceedings of the 2018 Conference of the North
  American Chapter of the Association for Computational Linguistics: Human
  Language Technologies, Volume 2 (Short Papers)}, pp.~464--468, 2018.

\bibitem{ntm}
A.~Graves, G.~Wayne, and I.~Danihelka, ``{Neural Turing Machines},'' 2014.

\bibitem{stackrnn}
A.~Joulin and T.~Mikolov, ``{Inferring Algorithmic Patterns with
  Stack-Augmented Recurrent Nets},'' in {\em Proceedings of the 28th
  International Conference on Neural Information Processing Systems - Volume
  1}, NIPS’15, (Cambridge, MA, USA), p.~190–198, MIT Press, 2015.

\bibitem{nueralgpu}
Łukasz Kaiser and I.~Sutskever, ``{Neural GPUs Learn Algorithms},'' 2015.

\bibitem{dnc}
A.~Graves, G.~Wayne, M.~Reynolds, T.~Harley, I.~Danihelka,
  A.~Grabska-Barwi{\'n}ska, S.~G. Colmenarejo, E.~Grefenstette, T.~Ramalho,
  J.~Agapiou, {\em et~al.}, ``{Hybrid Computing Using a Neural Network with
  Dynamic External Memory},'' {\em Nature}, vol.~538, no.~7626, pp.~471--476,
  2016.

\bibitem{rtransformer}
Z.~Wang, Y.~Ma, Z.~Liu, and J.~Tang, ``{R-transformer: Recurrent Neural Network
  Enhanced Transformer},'' {\em arXiv preprint arXiv:1907.05572}, 2019.

\bibitem{arn}
J.~Hao, X.~Wang, B.~Yang, L.~Wang, J.~Zhang, and Z.~Tu, ``{Modeling Recurrence
  for Transformer},'' in {\em Proceedings of the 2019 Conference of the North
  {A}merican Chapter of the Association for Computational Linguistics: Human
  Language Technologies, Volume 1 (Long and Short Papers)}, (Minneapolis,
  Minnesota), pp.~1198--1207, Association for Computational Linguistics, June
  2019.

\bibitem{highway}
T.-R. Chiang, C.-W. Huang, S.-Y. Su, and Y.-N. Chen, ``{Learning Multi-Level
  Information for Dialogue Response Selection by Highway Recurrent
  Transformer},'' {\em Computer Speech and Language}, vol.~63, p.~101073, 2020.

\bibitem{rtn}
S.~Kim, S.~Lin, S.~R. JEON, D.~Min, and K.~Sohn, ``{Recurrent Transformer
  Networks for Semantic Correspondence},'' in {\em Advances in Neural
  Information Processing Systems}, pp.~6126--6136, 2018.

\bibitem{tft}
B.~Lim, S.~{\"{O}}. Arik, N.~Loeff, and T.~Pfister, ``{Temporal Fusion
  Transformers for Interpretable Multi-horizon Time Series Forecasting},'' {\em
  CoRR}, vol.~abs/1912.09363, 2019.

\bibitem{recursivemind}
M.~C. Corballis, {\em {The recursive mind: The origins of human language,
  thought, and civilization-updated edition}}.
\newblock Princeton University Press, 2014.

\bibitem{fairseq}
M.~Ott, S.~Edunov, A.~Baevski, A.~Fan, S.~Gross, N.~Ng, D.~Grangier, and
  M.~Auli, ``{fairseq: A fast, extensible toolkit for sequence modeling},''
  {\em arXiv preprint arXiv:1904.01038}, 2019.

\bibitem{huggingface}
``{Hugging Face Transformers}.''
  \url{https://github.com/huggingface/transformers}, 2020.

\bibitem{character-level-transformer}
R.~Al-Rfou, D.~Choe, N.~Constant, M.~Guo, and L.~Jones, ``Character-level
  language modeling with deeper self-attention,'' in {\em Proceedings of the
  AAAI Conference on Artificial Intelligence}, vol.~33, pp.~3159--3166, 2019.

\bibitem{liang2017combining}
D.~Liang, W.~Xu, and Y.~Zhao, ``Combining word-level and character-level
  representations for relation classification of informal text,'' in {\em
  Proceedings of the 2nd Workshop on Representation Learning for NLP},
  pp.~43--47, 2017.

\bibitem{socialimpact}
D.~Hovy and S.~L. Spruit, ``{The social impact of natural language
  processing},'' in {\em 54th Annual Meeting of the Association for
  Computational Linguistics, ACL 2016 - Short Papers}, pp.~591--598, 2016.

\end{thebibliography}
